\documentclass[letterpaper]{article} %
\usepackage{aaai23}  %
\usepackage{times}  %
\usepackage{helvet}  %
\usepackage{courier}  %
\usepackage[hyphens]{url}  %
\usepackage{graphicx} %
\usepackage{natbib}  %
\usepackage{caption} %
\usepackage{algorithm}
\usepackage{algorithmic}

\usepackage{amsmath,amsfonts,bm}

\def\eqref#1{equation~\ref{#1}}

\def\1{\bm{1}}

\DeclareMathAlphabet{\mathsfit}{\encodingdefault}{\sfdefault}{m}{sl}
\SetMathAlphabet{\mathsfit}{bold}{\encodingdefault}{\sfdefault}{bx}{n}

\def\gE{{\mathcal{E}}}

\newcommand{\E}{\mathbb{E}}

\DeclareMathOperator*{\argmax}{arg\,max}

\usepackage{newfloat}
\usepackage{listings}
\DeclareCaptionStyle{ruled}{labelfont=normalfont,labelsep=colon,strut=off} %
\floatstyle{ruled}
\newfloat{listing}{tb}{lst}{}
\floatname{listing}{Listing}
\usepackage[utf8]{inputenc} %
\usepackage[T1]{fontenc}    %
\usepackage{url}            %
\usepackage{booktabs}       %
\usepackage{amsfonts}       %
\usepackage{nicefrac}       %
\usepackage{microtype}      %
\usepackage{xcolor}         %

\usepackage[textsize=tiny]{todonotes}

\usepackage{comment}
\RequirePackage{algorithm}
\RequirePackage{algorithmic}
\usepackage{blindtext}
\usepackage{graphicx}
\usepackage{multirow}
\usepackage{array}
\usepackage{caption}
\usepackage{subcaption}
\usepackage{amsmath}
\usepackage{amssymb}
\usepackage{amsthm}
\usepackage{mathtools}
\usepackage{adjustbox}
\usepackage{color}

\newcommand{\qedwhite}{\hfill \ensuremath{\Box}}

\def\ie{\textit{i.e.,~}}
\def\eg{\textit{e.g.,~}}

\def\wrt{\textit{w.r.t.~}}

\newtheorem{theorem}{Theorem}[section]
\newtheorem{prop}[theorem]{Proposition}

\newtheorem{remark}[theorem]{Remark}

\title{On the Connection between Invariant Learning and Adversarial Training for Out-of-Distribution Generalization}
\author{
    Shiji Xin$^{1,4}$, Yifei Wang$^2$, Jingtong Su$^3$, Yisen Wang$^{1,5}$\thanks{Corresponding Author: Yisen Wang (yisen.wang@pku.edu.cn).}
}
\affiliations{
    \textsuperscript{1} Key Lab.~of Machine Perception (MoE),\\
    School of Intelligence Science and Technology, Peking University \\
    \textsuperscript{2} School of Mathematical Sciences, Peking University \\
    \textsuperscript{3} Center for Data Science, New York University \\
    \textsuperscript{4} School of EECS, Peking University \\
    \textsuperscript{5} Institute for Artificial Intelligence, Peking University 
}

\usepackage{bibentry}

\begin{document}

\maketitle

\begin{abstract}
Despite impressive success in many tasks, deep learning models are shown to rely on spurious features, which will catastrophically fail when generalized to out-of-distribution (OOD) data. Invariant Risk Minimization (IRM) is proposed to alleviate this issue by extracting domain-invariant features for OOD generalization. Nevertheless, recent work shows that IRM is only effective for a certain type of distribution shift (\eg correlation shift) while it fails for other cases (\eg diversity shift). Meanwhile, another thread of method, Adversarial Training (AT), has shown better domain transfer performance, suggesting that it has the potential to be an effective candidate for extracting domain-invariant features. This paper investigates this possibility by exploring the similarity between the IRM and AT objectives. Inspired by this connection, we propose Domain-wise Adversarial Training (DAT), an AT-inspired method for alleviating distribution shift by domain-specific perturbations. Extensive experiments show that our proposed DAT can effectively remove domain-varying features and improve OOD generalization under both correlation shift and diversity shift.
\end{abstract}

\section{Introduction}

\label{sec:intro}
Modern deep learning techniques have achieved remarkable success in many tasks \cite{he2016deep,wang2017residual,brown2020language}. However, deep models will suffer catastrophic performance degradation under some scenarios, as they tend to exploit spurious correlations in the training data \cite{beery2018recognition}. One of those representative scenarios is the Out-of-Distribution (OOD) generalization,
where the trained model is expected to perform well at the test time, even when the training and testing data come from different distributions \cite{zhang2021can}.
Another representative scenario in which deep models are unstable is the adversarial example. Researchers have found that deep models are quite brittle, as one can inject imperceptible perturbations into the input and cause the model to make wrong predictions with extremely high confidence \cite{szegedy2013intriguing}. 

\begin{figure}[!t]
    \centering
    \includegraphics[width=0.9\linewidth]{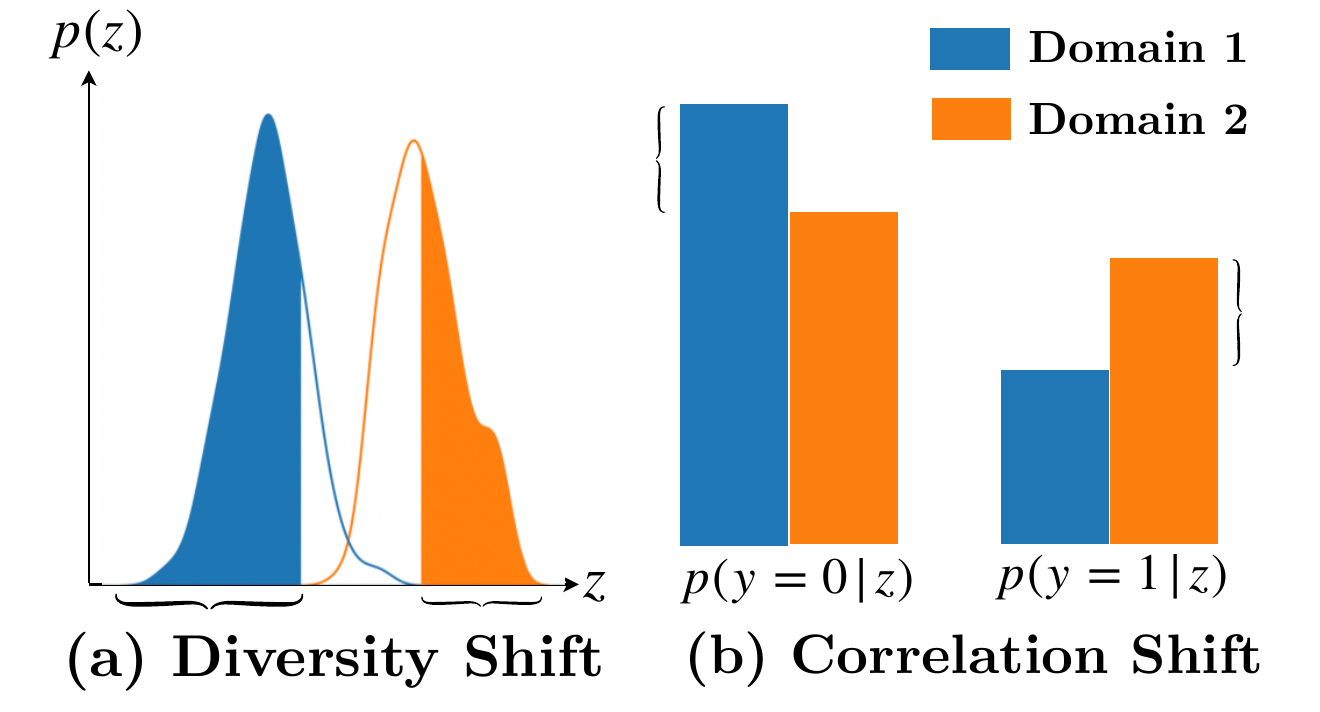}
    \caption{An illustrative example of the two kinds of distribution shifts. The curly brackets enclose the distribution shift between the environments. Here $z$ stands for the spurious feature, and $y$ stands for label class.}
    \label{fig:shifts}
\end{figure}

These two issues have some similarities to each other. They both arise because deep networks do not learn the essential causal associations (or intrinsic features). Nevertheless, in their corresponding fields, different approaches have been proposed. A large class of methods called Invariant Causal Prediction (ICP) \cite{peters2016causal} is proposed for OOD generalization. Among them, Invariant Risk Minimization (IRM) \cite{arjovsky2019invariant} attracts significant attention, which intends to extract invariant features across different data distributions and expects the model to ignore information related to the environment\footnote{The terminologies of ``domain'', ``environment'', and ``distribution'' are often used interchangeably in current literature.}. While for adversarial robustness against adversarial examples, Adversarial Training (AT) \cite{madry2018towards,wang2019dynamic} is the most effective approach at the current stage \cite{athalye2018obfuscated}. It trains a model on adversarial examples generated by injecting %
perturbations optimized for each image into natural examples.
Several recent works have explored the relationship between AT and OOD \cite{volpi2018generalizing, Shankar2018GeneralizingAD, Yi2021ImprovedOG}, but rarely focus on the typical domain generalization setting considered by IRM. 
Therefore, the two fields still seem rather independent. In this paper, we are going to explore their potential relationships.

Although IRM and its variants are promising on certain tasks, \eg CMNIST \cite{arjovsky2019invariant}, recent studies \cite{gulrajani2020search} show that on a large-scale controlled experiment on OOD generalization, all these methods fail to exceed the simplest \emph{i.i.d.}  baseline, \ie Empirical Risk Minimization (ERM). 
Further, two kinds of distribution shifts in benchmark datasets are identified \cite{ye2021ood}, \ie, diversity shift and correlation shift, shown in Figure \ref{fig:shifts}. Diversity shift refers to the shift of the distribution support of spurious feature $z$, for example, the style of the images changed from cartoon to sketch on object classification task. In contrast, correlation shift refers to the change in conditional probability (posterior distribution) of label $y$ given spurious feature $z$ on the same support, \eg, the color in the CMNIST dataset \citep{ahuja2020invariant}. They found that an algorithm that performs well on one kind of distribution shift tends to perform poorly on the other one \cite{ye2021ood}.

Thus, we need to seek better alternatives for OOD generalization, while AT seems to be a promising candidate from both theoretical and empirical aspects. Theoretically, by learning invariance \wrt local input perturbations \cite{wu2020adversarial}, AT can be regarded as Distributionally Robust Optimization (DRO) \cite{sinha2018certifying, volpi2018generalizing, rahimian2019distributionally, duchi2021statistics} over the $\ell_p$-bounded distributional shift. Thus, AT could reliably extract robust features, \eg the shape of the object, from the input \cite{ren2021unified}. Empirically, several recent works show that AT has better domain transferability than ERM \cite{salman2020adversarially, Yi2021ImprovedOG}. These findings naturally lead to the following questions: 
\begin{center}
    \emph{Is AT related to IRM? If so, is AT helpful for OOD generalization?}
\end{center}

In this paper, we take a further step to answer these intriguing questions. We first reveal the connections between IRM and AT, and find that IRM can be regarded as an instance-reweighted version of Domain-wise Adversarial Training (DAT), a new version of adversarial training that we propose for multi-source domain generalization.
Inspired by this connection, we further explore how DAT performs on OOD data. We first notice that DAT is suitable for solving domain generalization problems, as it can effectively remove relatively static background information with domain-wise perturbations. We further verify this intuition on both synthetic tasks and real-world datasets, where DAT shows clear advantages over ERM. Finally, we conduct extensive experiments on benchmark datasets and show that our DAT can consistently outperform ERM on tasks dominated by both correlation shift and diversity shift.

We summarize our contributions as follows:
\begin{itemize}
    \item We theoretically derive the connection between IRM and AT. Based on this connection, we develop a new kind of adversarial training, Domain-wise Adversarial Training (DAT), for domain generalization.
    \item We analyze how DAT is beneficial for learning invariant features and verify our hypothesis through synthetic data and real-world datasets.
    \item Experiments on benchmark datasets show that DAT not only performs better than ERM under correlation shift like IRM but also outperforms ERM under diversity shift like (sample-wise) AT.
\end{itemize}

\section{Related Works}

\label{sec:related}
\paragraph{IRM and Its Variants.} %
Invariant Risk Minimization (IRM) \cite{arjovsky2019invariant} develops a paradigm to extract causal (invariant) features and find the optimal invariant classifier on top of several given training environments. 
The work of \citet{kamath2021does} reveals the gap between IRM and IRMv1, showing that even in a simple model that echos the idea of the original IRM objective, IRMv1 can fail catastrophically. \citet{rosenfeld2020risks} prove that when the number of training environments is not large enough, IRMv1 can face the risk of using environmental features.

\paragraph{AT and Its Variants.} \citet{szegedy2013intriguing} reported that one can inject imperceptible perturbations to fool deep models. Among the proposed defenses, Adversarial Training (AT) \cite{goodfellow2014explaining, madry2018towards,wang2020improving,wang2022self} is the promising and representative approach to training robust models \cite{athalye2018obfuscated}. Recently, \citet{salman2020adversarially} showed that adversarially learned features could transfer better than standardly trained models, while various works \cite{volpi2018generalizing, sinha2018certifying, Shankar2018GeneralizingAD, Ford2019AdversarialEA, Qiao2020LearningTL, Yi2021ImprovedOG, gokhale2021} adopt sample-wise adversarial training or adversarial data augmentation to improve OOD robustness. However, most discussions are limited to distributional robustness \wrt Wasserstein distance. The small perturbations used in AT make it less practical to account for real-world OOD scenarios (\eg correlation to backgrounds), thus \citet{wang2022improving} incorporated low-rank structured priors into AT for this kind of large data distribution shifts. 
Moreover, previous work shows that there seems to be a trade-off between the two distribution shifts: an algorithm that performs well on one task tends to perform poorly on the other \cite{ye2021ood}. %
Instead, in this work, our proposed method can achieve fair performance on both correlation shift and diversity shift tasks.

\section{Relationship Between IRM and AT Variants}
\paragraph{Notation.}
Let $\Phi :\mathcal X \subset R^n\to R^d$ denote the representation of a $\theta$-parameterized piecewise linear classifier, \ie $\Phi(\cdot)=\phi^{L}(W^{L}\phi^{L-1}(\dots)+b^{L-1})+b^L$, where $\phi^L$ is the activation function, and $W^L$, $b^L$ denote the layer-wise weight matrix and bias vector, collectively denoted by $\theta$. Furthermore, let $\beta$ be the linear classifier on the top, and let the network output be $\beta\cdot\Phi(x)=\beta^\top\Phi(x)$. Let $\ell(\hat y,y)=-\log\sigma(y\hat y)$ be the sample logistic loss. We consider a two-class $(y=\pm 1)$ classification setting with output dimension $d=1$, and our discussion can be easily extended to general cases.

\paragraph{ERM.}
The traditional Empirical Risk Minimization (ERM) algorithm optimizes over the loss on i.i.d. data, \ie
\begin{equation}
    \begin{aligned}
  \min_{\beta,\Phi} R(\beta\cdot\Phi), \ \mathrm{where}\ R(\beta\cdot \Phi)=\mathbb E_{(x,y)\sim D} \ell(\beta^\top \Phi(x),y).
    \end{aligned}
\end{equation}
In the OOD generalization problem, one faces a set of (training) \textit{environments} $\mathcal E$, where each environment $e\in\mathcal E$ corresponds to a unique data distribution $D_e$. %
When facing multiple environments, the ERM objective simply mixes the data together and takes the form
\begin{equation}
       (\operatorname{ERM})\quad
       \min_{\beta,\Phi} \sum_e R^e(\beta\cdot\Phi),
\end{equation}
where $R^e(\beta\cdot \Phi)=\mathbb E_{(x,y)\sim D_e} \ell(\beta^\top \Phi(x),y)$.

\paragraph{IRM and IRMv1.}
Instead of simply mixing the data together, IRM seeks to learn an \emph{invariant} representation $\Phi$ such that the objective can be minimized with the same classifier $\beta$ in all training domains. Formally, we have
\begin{equation}
(\operatorname{IRM})\quad
\begin{aligned}
    \min_{\beta,\Phi}&\sum_{e\in\mathcal E}R^e(\beta\cdot \Phi)%
    \\
    \mathrm{s.t.}\ &\beta\in\arg\min_{\bar \beta}R^e(\bar \beta\cdot \Phi),\forall e\in\mathcal E.
\end{aligned}
\end{equation}
Since this bilevel optimization problem is difficult to solve, the practical version IRMv1 is formulated as regularized ERM, where the gradient penalty is calculated \wrt a dummy variable $w$:
\begin{equation} \label{IRMv1Obj}
(\operatorname{IRMv1}) 
\min_{\beta,\Phi}\sum_{e\in\mathcal E}\!\big[R^e(\beta\cdot \Phi)+\!\lambda\!\underbrace{||\nabla_{w|w=1.0}R^e(w\cdot(\beta\cdot \Phi))||^2}_{\mathrm{Penalty_{IRM}}}\big].
\end{equation}

\paragraph{AT.} Adversarial Training instead replaces ERM with a minimax objective,

\begin{equation}
\min_{\beta,\Phi} R^{\rm AT}(\beta\cdot\Phi) = \min_{\beta,\Phi}  \mathbb E_{(x,y)\sim D}\max_{||\delta_x||_p\le \varepsilon} \ell(\beta^\top \Phi(x+\delta_x),y),
\end{equation}
where one maximizes inner loss by injecting \emph{sample-wise} perturbations $\delta_x$ and solve the outer minimization \wrt parameters $\beta,\Phi$ on the perturbed sample $(x+\delta_x, y)$. Typically, the perturbation is constrained within an $\ell_p$ ball with radius $\varepsilon$. In this way, AT can learn models that are robust to $\ell_p$ perturbations.

\subsection{Relating AT to IRM}

As shown above, it seems that IRM and AT are two distinct learning paradigms, while, in fact, we can show that IRM is closely related to a certain kind of adversarial training. To see this, we first notice that AT can be rephrased into a regularized ERM loss with a penalty on \emph{sample-wise} robustness through linearization:
\begin{equation}
\label{eq:at}
\begin{aligned}
&R^{\rm AT}(\beta\cdot\Phi) = \mathbb E_{(x,y)\sim D}\max_{||\delta_x||\le \varepsilon} \ell(\beta^\top \Phi(x+\delta_x),y) \\
=\ &\mathbb E_{(x,y)\sim D}\big[\ell(\beta^\top \Phi(x),y) +\\
&\max_{||\delta_x||\le \varepsilon}\left( \ell(\beta^\top \Phi(x+\delta_x),y)-\ell(\beta^\top \Phi(x),y)\right)\big]\\
\approx\ & R(\beta\cdot\Phi) + \varepsilon\underbrace{\E_{(x,y)\sim D}\left\|\nabla_{x}\ell(\beta^\top\Phi(x),y)\right\|}_{\rm Penalty_{AT}},
\end{aligned}
\end{equation}
which resembles the gradient penalty adopted in IRMv1. One main difference is that AT's penalty is calculated \wrt sample-wise gradients, while IRM's penalty \wrt the population loss. This difference motivates us to adopt a population-level perturbation $\delta$ instead.

\paragraph{Proposed DAT.} Inspired by the above connection, we propose Domain-wise Adversarial Training (DAT), which adopts a \emph{domain-wise} perturbation $\delta_e$ for each training domain $e\in\gE$. Formally, we have 
\begin{equation}
\begin{aligned}
    \min_{\beta,\Phi}\ & \sum_{e\in\mathcal E} \mathbb E_{(x,y)\sim D_e} \ell\left(\beta^\top\Phi(x+\delta_e)),y\right)\\
    \mathrm{s.t.}\ &\delta_e\in\argmax_{\|\delta\|\le \varepsilon} \mathbb E_{(x,y)\sim D_e}\ell(\beta^\top\Phi(x+ \delta),y),\forall e \in\mathcal E.
\end{aligned}
\label{eq:dat}
\end{equation}
where the perturbation $\delta_e$ is defined at the distribution level.

In practice, we solve the above problem by alternating updates of model parameters $\beta,\Phi$ and perturbations $\delta_{e}$.
Specifically, for each mini-batch $B_e$ sampled from domain $\mathcal{D}_e$, we update $\delta_e$ with $B_e$ using gradient ascent to find the best adversarial perturbations. The adversarial samples are then used to train the model. A detailed description is in Algorithm \ref{alg:dat}.

\begin{algorithm}[ht]
\caption{Domain-wise Adversarial Training}\label{alg:dat}
\begin{algorithmic}
\REQUIRE Dataset of multiple environments $D_e, e\in \mathcal{E}$,\\ desired $l_p$ norm of the perturbation $\varepsilon$,\\ perturbation step size $\alpha$
\ENSURE Model $(\Phi, \beta)$
\STATE Randomly initiate $\theta$, perturbation $\delta_e, \forall e\in \mathcal{E}$
\FOR{iterations $=1, 2, 3, \ldots$}
	\FOR{each environment $e$}
	    \STATE 1. Sample batch $B_e$ from environment $e$
        \STATE 2. Update the perturbation $\delta_e$ using one-step\\ \quad gradient ascent with step size $\alpha$
        \STATE 3. Project the perturbation $\delta_e$ to a $\ell_p$ ball of radius $\varepsilon$ 
        \STATE 4. Generate adversarial examples \\\quad$x_{adv} \gets x+\delta_e, \forall x\in B_e$
        \STATE 5. Update $\Phi$ and $\beta$ with gradient descent on $x_{adv}$
    \ENDFOR
\ENDFOR
\end{algorithmic}
\end{algorithm}

In the setting of single-source domain generalization, where only one training domain is provided, DAT degenerates into universal adversarial training (UAT) \cite{moosavi2017universal}, where a single perturbation is provided for the entire training distribution.

\subsection{Connection Between IRM and DAT}

Here, we establish a formal connection between IRM and DAT. To begin with, we note that DAT can also be reformulated as a regularized ERM in the multi-domain scenario.
\begin{equation}
\begin{aligned}
&R^{\rm DAT}(\beta\cdot\Phi) =\sum_{e\in\gE} \max_{||\delta_e||\le \varepsilon}\mathbb E_{(x,y)\sim D_e} \ell(\beta^\top \Phi(x+\delta_e),y) \\
=\ & \sum_{e\in\gE}\big[\E_{(x,y)\sim D_e}\ell(\beta^\top \Phi(x),y)+ \\
&\max_{||\delta_e||\le \varepsilon}\mathbb E_{(x,y)\sim D_e}\left( \ell(\beta^\top \Phi(x+\delta_e),y)-\ell(\beta^\top \Phi(x),y)\right)\big]\\
\approx\ & \sum_{e\in\gE}\big[R^e(\beta\cdot\Phi) + \varepsilon\underbrace{\left\| \mathbb E_{(x,y)\sim D_e}\nabla_x \ell(\beta^\top\Phi(x),y)\right\|}_{\rm Penalty_{DAT}}\big].
\end{aligned}
\label{eq:dat2}
\end{equation}

Based on this reformulation, we can show that there exists an intrinsic relationship between DAT and IRM as in the following proposition:

\begin{prop}
\label{prop2}
Consider each $D_e$ as the corresponding distribution of a particular training domain $e$. For any $\beta\cdot\Phi$ as a deep network with any activation function, the penalty term of IRMv1, $\mathrm{Penalty_{IRM}}$ (Eq.~\ref{IRMv1Obj}), could be expressed as the square of a reweighted version of the penalty term of the above approximate target, $\mathrm{Penalty_{DAT}}$(Eq.~\ref{eq:dat2}), on each environment $e$ with coefficients related to 
the distribution $D_e$, which could be stated as follows:
\begin{align}
    \label{irm}
    \mathrm{Penalty_{IRM}} 
    &= \left\|\mathbb{E}_{D_{e}} [L_x x + \tilde{B}_x]\right\|^2\\ %
    \mathrm{Penalty_{DAT}} &= \left\|\mathbb{E}_{D_{e}} L_x\right\|
\end{align}
where $L_x = (1-\sigma(y\beta^\top\Phi_x x))y\beta^\top\Phi_x$ and $\tilde{B}_{x}=\left(1-\sigma\left(y \beta^{\top} \Phi_{x} x\right)\right) y \beta^{\top} B_{x}$. $B_x$ denotes the collection of constants introduced by bias terms in neural network layers.
\end{prop}

If we consider the extreme case where each domain only contains one sample, we can see that DAT degenerates into AT as a special case. The equivalence between IRMv1 and (linearized) DAT in this setting can be shown as follows, which extends the similarity between IRM (Eq. \ref{IRMv1Obj}) and AT (Eq. \ref{eq:at}).

\begin{remark}[Equivalence under Single-sample Environments]
\label{remark}
    When the environments degenerate into a single data point, we have the following relationship:
If $\varepsilon$ is sufficiently small, then for $\beta\cdot\Phi$ as a deep network with any activation function,
the penalty term of IRMv1 (Eq. \ref{IRMv1Obj}) on each sample and the square of the maximization term of the linearized version of Eq.~\ref{eq:dat} (LDAT, obtained by the first-order approximation of DAT) 
\begin{equation}
    \mathrm{Penalty_{LDAT}} = \left\langle\nabla_{x} \ell\left(\beta^{T} \Phi(x), y\right), \pm\hat\delta_x\right\rangle
\end{equation}
on each sample with perturbation\ $\hat\delta_x = \pm\varepsilon x$ only differ by a fixed multiple $\varepsilon^2$ and a bias term $B_x$, which is formally stated as
\begin{equation}
\begin{aligned}
\mathrm{Penalty^{2}_{LDAT}}
&= [\langle \nabla_{x} \ell\left(\beta^{T} \Phi(x), y\right),\pm \varepsilon x\rangle]^2 \\
&=\varepsilon^2 (1-\sigma(\beta^\top\Phi(x)))^2 \left\|\beta^\top \Phi_x x\right\|^2\\
&= \varepsilon^{2}\cdot\mathrm{Penalty^\prime_{IRM}},
\end{aligned}
\end{equation}
where $\mathrm{Penalty^\prime_{IRM}}=(1-\sigma(y\beta^\top\Phi(x)))^2 \left\|\beta^\top(\Phi_x x)\right\|^2$, $\mathrm{Penalty_{IRM}}=(1-\sigma(y\beta^\top\Phi(x)))^2 \left\|\beta^\top(\Phi_x x+B_x)\right\|^2$.
\end{remark}
The proofs of Proposition~\ref{prop2} and Remark~\ref{remark} can be found in Appendix~\ref{proof}. The above connection between DAT and IRM highlights that our DAT is potentially helpful in addressing OOD problems. 

\section{Empirical Investigation on Domain-wise Adversarial Training}
\label{4}

In this section, we further explore how domain-wise perturbations could help alleviate distribution shifts in real-world scenarios. In particular, we noticed that domain-wise perturbations could effectively remove the domain-varying \emph{background information}, which usually corresponds to spurious features for image classification tasks. We empirically verify this property by applying DAT to a well-designed OOD task based on background shift.

\begin{figure}[]
\centering
\begin{subfigure}[]{0.48\textwidth}
    \includegraphics[width=\linewidth]{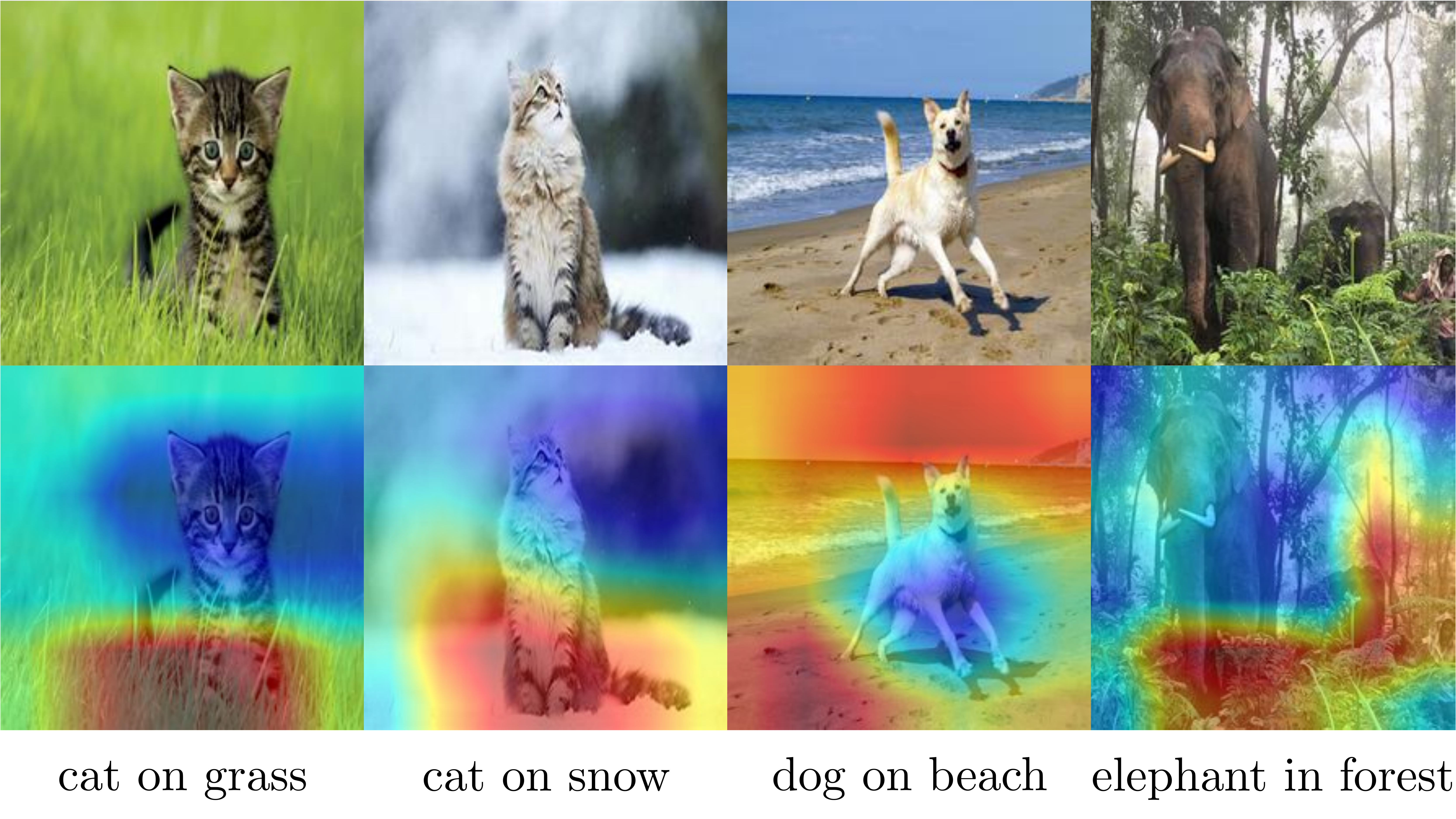}
    \vspace{-0.2 in}
    \caption{Images and heatmaps of ERM from NICO.} 
    \label{fig:NICO}
\end{subfigure}
\hfill
\begin{subfigure}[]{0.45\textwidth}
    \vspace{0.2 in}
    \includegraphics[width=\linewidth]{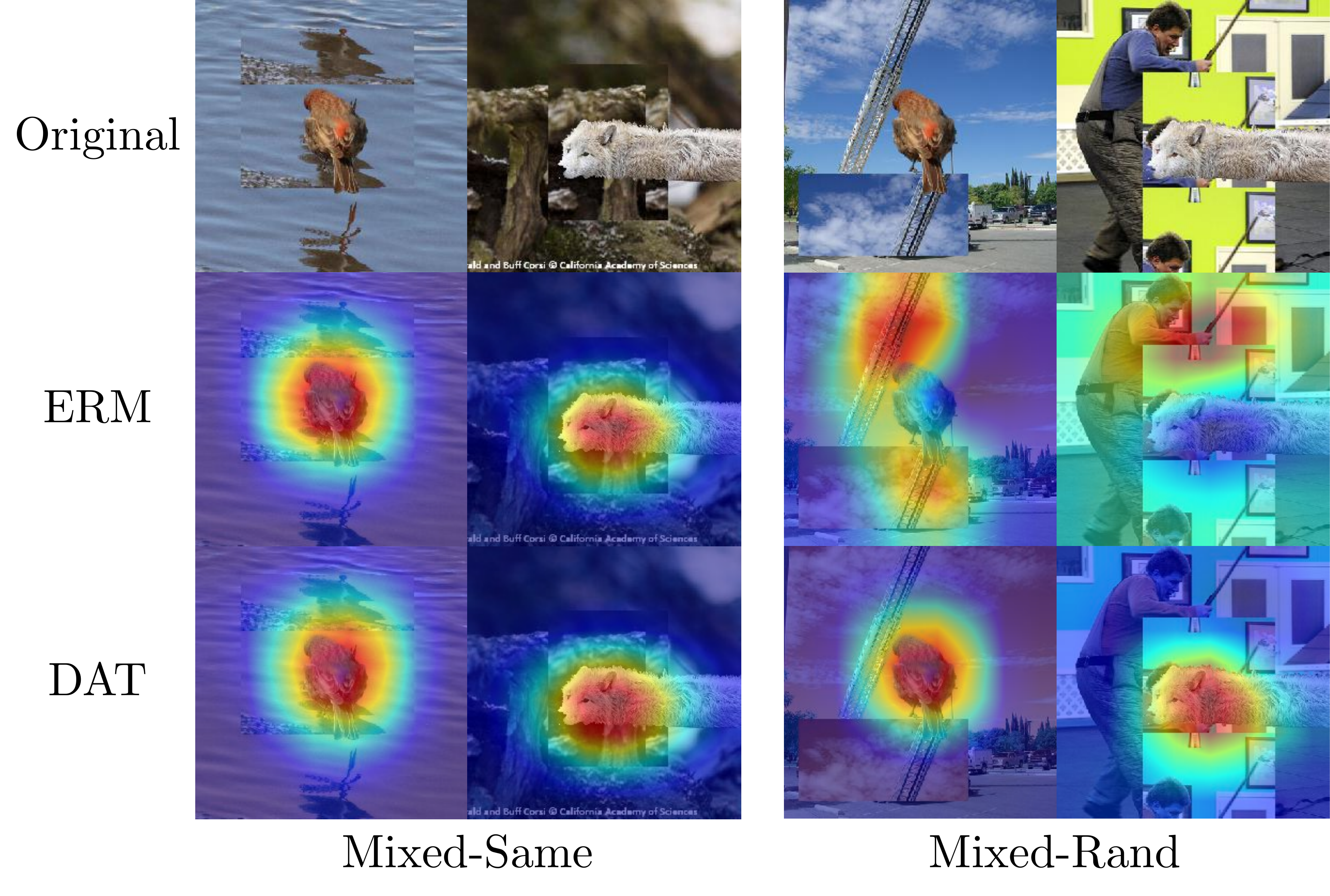}
    \vspace{-0.2 in}
    \caption{Images and heatmaps of ERM and DAT from Mixed-Same and Mixed-Rand.}
    \label{fig:mixed}
\end{subfigure}
\caption{Images from NICO and Mixed datasets and corresponding heatmaps of the models. The red regions in the heatmaps correspond to the model's focus during prediction. Heatmaps demonstrate the effectiveness of DAT in attending to interested objects rather than uncorrelated background information compared to ERM.}
\end{figure}

\subsection{Comparing ERM to DAT for Background Removal}
\paragraph{ERM Learns Spurious Background Features.} Our understanding of DAT is based on the idea that an image is composed of a foreground object and a corresponding background, and typically the object is the invariant feature, while the background is only spuriously correlated with the labels.
However, models that rely on spurious background information will easily fail when encountering images from a different domain.
This phenomenon is also empirically verified by \citet{xiao2020noise}, who find that models trained on an ImageNet-like dataset with ERM require image backgrounds to correctly classify large portions of test sets. 
These findings point out the limitations of ERM and motivate us to find a solution that could effectively learn a background-invariant classifier.

\begin{table*}[ht]
$$
\begin{array}{lccccc}
\toprule
\multirow{2}{*}{Algorithm} & \multicolumn{2}{c}\text{Correlation shift\qquad} &\multicolumn{2}{c}\text{Diversity shift\ \ \qquad}\\
\cmidrule(lr){2-3} \cmidrule(lr){4-5}
& \text{CMNIST}
& \text{NICO } & \text{ PACS } & \text{ TerraInc } & \\
\midrule
\text{ERM (Baseline)} & 58.5 \pm 0.3 & 72.1 \pm 1.6 & 81.5\pm 0.0 & 42.6 \pm 0.9\\
\midrule
\text{VREx \cite{krueger2020out}} & 56.3 \pm 1.9 & 71.5 \pm 2.3 & 81.8 \pm 0.1 & 40.7 \pm 0.7\\
\text{GroupDRO \cite{sagawa2019distributionally}} & 61.2 \pm 0.6 & 71.0 \pm 0.4 & 80.4 \pm 0.3 & 36.8 \pm 1.1\\
\text{IRM \cite{ahuja2020invariant}} & \mathbf{70.2 \pm 0.2} & \mathbf{73.3 \pm 2.1} & 81.1 \pm 0.3 & 42.0 \pm 1.8 \\
\text{ARM \cite{zhang2020adaptive} } & 63.2 \pm 0.7 & 67.3 \pm 0.2 & 81.0 \pm 0.4 & 39.4 \pm 0.7\\
\text{RSC \cite{huang2020self} } & 58.5 \pm 0.5 & \mathbf{74.3 \pm 1.9} & \mathbf{82.8 \pm 0.4} & \mathbf{43.6 \pm 0.5}\\
\text{DANN \cite{ganin2016domain}} & 58.3 \pm 0.2 & 69.4 \pm 1.7 & 81.1 \pm 0.4 & 39.5 \pm 0.2\\
\text{MMD \cite{li2018domain}} & 63.4 \pm 0.7 & 68.9 \pm 1.2 & 81.7 \pm 0.2 & 38.3 \pm 0.4 \\
\text{MTL \cite{blanchard2017domain} } & 57.6 \pm 0.3 & 70.6 \pm 0.8 & 81.2 \pm 0.4 & 38.9 \pm 0.6\\
\text{MLDG \cite{li2018learning}} & 58.4 \pm 0.2 & 66.6 \pm 2.4 & 73.0 \pm 0.4& 27.3 \pm 2.0\\
\text{SagNet \cite{nam2019reducing} } & 58.2 \pm 0.3 & 69.8 \pm 0.7 & 81.6 \pm 0.4 & 42.3 \pm 0.7\\
\text{CORAL \cite{sun2016deep}} & 57.6 \pm 0.5 & 70.8 \pm 1.0 & 81.6 \pm 0.6 & 38.3 \pm 0.7 \\
\text{Mixup \cite{yan2020improve}} & 58.4 \pm 0.2 & 72.5 \pm 1.5 & 79.8 \pm 0.6 & 39.8 \pm 0.3\\
\midrule
\text{AT (sample-wise) \cite{goodfellow2014explaining}} & 57.9 \pm 0.4 & 70.5 \pm 0.7 & \mathbf{82.0 \pm 0.2}
& 42.6 \pm 0.3\\
\text{UAT \cite{shafahi2020universal}} & 58.7 \pm 2.3 & 69.1 \pm 1.2 & 80.7 \pm 0.4 & 41.9 \pm 1.8\\
\text{WRM \cite{sinha2018certifying}} & 57.9 \pm 3.3 & 68.2 \pm 1.0 & 80.4 \pm 0.0 & 26.1 \pm 1.5\\
\text{ADA \cite{volpi2018generalizing}} & 56.3 \pm 0.4 & 69.5 \pm 1.9 & 80.2 \pm 0.2 & 41.2 \pm 0.7\\
\midrule
\text{DAT (our work)} & \mathbf{68.4 \pm 2.0} & 72.6 \pm 1.7 & \mathbf{82.0 \pm 0.1} & \mathbf{42.7 \pm 0.7}\\
\bottomrule
\end{array}
$$
\vspace{-0.15 in}
\caption{Test accuracy (\%) on four representative tasks for OOD generalization. According to the OOD-Bench \cite{ye2021ood}, two are dominated by correlation shift, CMNIST and NICO, and two are dominated by diversity shift, PACS and TerraInc. We highlight the top two results on each task.}
\label{table:bench}
\end{table*}

\paragraph{Removing Background Information with DAT.} Compared to the failure of ERM above, we notice that DAT can be applied to eliminate domain-wise background information with its domain-specific perturbations. 
Take the NICO images in Figure \ref{fig:NICO} as an example, where samples from the environment ``on grass'' have a common background dominated by green grass with low frequency, while the foreground object (\eg the cats) has complex and instance-specific shape and texture with much higher frequency. 
In fact, \citet{moosavi2017universal} show that a universal perturbation vector lies in a low-dimensional subspace, which fits the background statistics and could be used to eliminate the low-frequency background factor. 
Therefore, when our DAT is applied to these samples, the domain-wise perturbation will capture the common domain-specific background. Consequently, domain-wise AT will help remove the dependence on these spurious background features.

\subsection{Empirical Verification with Controlled Experiments}
\label{4.2}
We construct a synthetic OOD task to verify the above analysis by evaluating a classifier's dependence on background information. It is based on two datasets introduced by \citet{xiao2020noise}, Mixed-Same and Mixed-Rand, which are constructed from a subset of ImageNet images with the background of each image replaced by another background that is either of the same class (Mixed-Same) or a random class (Mixed-Rand). As they are perfect candidates for evaluating a classifier in terms of its background dependence, we construct a new OOD task by using Mixed-Same as the training domain and evaluating the learned classifier on Mixed-Rand as the test domain. If the classifier relies heavily on background information, it will perform poorly in the test domain, where objects and backgrounds are disentangled.
In particular, the experiment results show that ERM achieves a test accuracy of ${71.9\%}$, while DAT achieves ${72.6\%}$ in the test domain with random background, which means that DAT has a better generalization ability by effectively removing background information. Sample images from the dataset and the corresponding attention heatmaps are shown in Figure \ref{fig:mixed}, demonstrating that ERM may lose focus when the background correlation is broken while DAT does not. Details of the experiment are shown in Appendix \ref{mixed}.

\section{Experiments}

For the experiments, we follow the setting in \citet{ye2021ood} and evaluate the OOD generalization on both types of distribution shift: diversity shift and correlation shift. In particular, we select four representative tasks. For the correlation shift, we use CMNIST \cite{arjovsky2019invariant}, a synthetic dataset on digit classification with color as the spurious feature, and NICO \cite{he2020towards}, a real-world dataset on object classification with context as the spurious feature. Regarding diversity shift, we use PACS \cite{li2017deeper} and Terra Incognita \cite{beery2018recognition}, which are both datasets consisting of natural images with four domains. 
To ensure fair evaluation, we perform all of our experiments following the evaluation pipeline of DomainBed \cite{gulrajani2020search}. Specifically, we use the same dataset splitting and model selection strategy as in \citet{ye2021ood} for each task. For datasets except for NICO, one of the domains is used as the test domain. We train the models in each run, treating one of the domains as the test domain and the rest as training domains, then report the average accuracy of all domains. For NICO, the training, test, and evaluation domains are predefined. We train the models on training domains, evaluate them on the evaluation domain for model selection, and report their accuracy on the test domain. More details of the experimental settings and domain-split results can be found in Appendix \ref{dataset}.

When training models using DAT, we first perform standard data augmentation \cite{gulrajani2020search}, then proceed with the update of the perturbation and model parameters as shown in Algorithm \ref{alg:dat}, where the perturbed samples are clipped to the legal range after data augmentation. We use DAT with perturbation bounded by $\ell_2$-norm in our experiments. To test DAT on a wider range of tasks, we also carry out experiments on DomainBed \cite{gulrajani2020search}, the results are shown in Appendix \ref{domainbed}.

\subsection{Evaluation on Benchmark Datasets}
We compare our results with previous work, including vanilla ERM, invariance-based methods including IRM \cite{ahuja2020invariant}, robust optimization methods including GroupDRO \cite{sagawa2019distributionally}, distribution matching methods including MMD \cite{li2018domain} and CORAL \cite{sun2016deep}, a method based on domain classifier DANN \cite{ganin2016domain}, and various other algorithms.
The results of the CMNIST dataset are adopted from \citet{gulrajani2020search}, which is the average of three domains, while the results of the other datasets are adopted from \citet{ye2021ood}. In addition to that, we implement and test four AT-based algorithms, including sample-wise adversarial training AT \cite{goodfellow2014explaining}, universal adversarial training UAT \cite{shafahi2020universal}, WRM \cite{sinha2018certifying}, and adversarial data augmentation ADA \cite{volpi2018generalizing}.

From Table 1, we can see that DAT consistently outperforms ERM and achieves good performance under both diversity and correlation shifts. Specifically,
DAT achieves much better results than ERM on correlation-shift tasks (CMNIST and NICO) like IRM. Second, we can see that most domain generalization algorithms at the moment cannot surpass ERM on tasks dominated by diversity shifts. Although RSC has great performance on these tasks, it performs much worse under correlation shifts. However, DAT consistently outperforms ERM and could account for both kinds of shifts. 
Furthermore, compared to other AT-based algorithms (\ie sample-wise AT, UAT, WRM, and ADA), DAT has a fair performance by considering a domain-wise perturbation that removes domain-varying spurious features. The results demonstrate the effectiveness of DAT in dealing with domain discrepancy.

\subsection{Extension to Single Domain Generalization}
As discussed in Section \ref{4}, DAT can reduce the influence of background even when no domain labels are given, which corresponds to the single-source domain generalization setting. We conduct experiments that strengthen this claim to see how DAT could help in the single-source domain generalization setting. The experimental setting can be found in Appendix \ref{ssdg}. 

The results on the four datasets used in Table \ref{table:bench} are shown in Table \ref{tab:single}. We can see that our DAT has fair performance on all four datasets compared to ERM, either under correlation shift or diversity shift. Although the difference is not as significant as in the multiple-domain setting, it shows that our DAT works for both single-domain and multi-domain generalization scenarios. In particular, its advantages are more significant with multiple domains, where the domain-wise perturbation mechanism is more effective.

We also test DAT on Digits, a common single-source domain generalization dataset consisting of five sub-datasets: MNIST \cite{lecun1998gradient}, MNIST-M \cite{ganin2015unsupervised}, SVHN \cite{Netzer2011ReadingDI}, SYN \cite{ganin2015unsupervised}, and USPS \cite{Denker1988NeuralNR}. We show the results in Table \ref{tab:dig}. The performance of ERM is adopted from \citet{Qiao2020LearningTL}. From these results, we can see that DAT retains its better generalization ability in the challenging single-domain generalization setting by outperforming ERM in all tasks and could be a promising alternative to ERM.

\begin{table}[H]
\centering
\resizebox{\columnwidth}{!}{
\begin{tabular}{cccccc}
\toprule
{ Algorithm} & { CMNIST} & { NICO} & { PACS} & { TerraInc} \\ \midrule
{ ERM} & { $45.8 \pm 0.7$} & { $63.2\pm 2.9$} & { $59.8\pm 1.6$} & { $27.3\pm 4.0$}  \\
{ DAT} & { $\mathbf{46.0\pm 2.0}$} & { $\mathbf{64.4\pm 0.7}$} & { $\mathbf{59.9\pm 2.4}$} & { $\mathbf{28.1\pm 4.0}$}  \\ \bottomrule
\end{tabular}
}
\vspace{-0.1 in}
\caption{Average test accuracy (\%) of ERM, DAT on four representative tasks (single-source domain generalization).}
\label{tab:single}
\end{table}

\begin{table}[H]
\centering
\resizebox{\columnwidth}{!}{
\begin{tabular}{cccccc}
\toprule
Algorithm    & SVHN  & MNIST-M & SYN & USPS  & Avg   \\ \midrule
ERM & 27.83 & 52.72& 39.65&	76.94 & 49.29 \\
DAT          & $\mathbf{28.2 \pm 1.0}$ & $\mathbf{55.8 \pm 1.4}$ & $\mathbf{43.1\pm 0.2}$ & $\mathbf{81.3 \pm 1.0}$ & $\mathbf{52.1}$ \\ \bottomrule
\end{tabular}
}
\vspace{-0.1 in}
\caption{Results of ERM and DAT on Digits dataset}
\label{tab:dig}
\end{table} 

\begin{table}[H]
\centering
\begin{tabular}{cll}
\toprule
Algorithm    & CMNIST         & NICO           \\ \midrule
UAT & $58.7 \pm 2.3$ & $69.1 \pm 1.2$\\
Ensemble UAT & $58.2 \pm 2.3$ & $60.8 \pm 0.2$ \\
DAT          & $\mathbf{68.4 \pm 2.0}$ & $\mathbf{72.6 \pm 1.7}$ \\ \bottomrule
\end{tabular}
\vspace{-0.1 in}
\caption{Results of UAT, Ensemble UAT, and DAT on CMNIST and NICO.}
\label{tab:euat}
\end{table}

Under this setting, DAT generates perturbations in the only training domain, thus degenerating into UAT. However, as shown in Table \ref{table:bench}, DAT performs much better in the multidomain setting, which shows the necessity of generating domain-specific perturbations. Furthermore, to show that models trained using DAT on multi-domains are not a trivial ensemble of models trained on single-domains, we train voting classifiers on CMNIST and NICO using UAT models trained separately on each domain. The results are shown in Table \ref{tab:euat}. We can see that neither UAT nor ensemble UAT has the same performance as DAT. This verifies the effectiveness of learning domain-wise perturbations using DAT.

\subsection{Analysis}
\label{analysis}
We conduct extensive experiments to better understand what our algorithm learns and how the magnitude of hyperparameters affects its performance.

\begin{figure}[!htbp]
    \centering
    \includegraphics[width=1\linewidth]{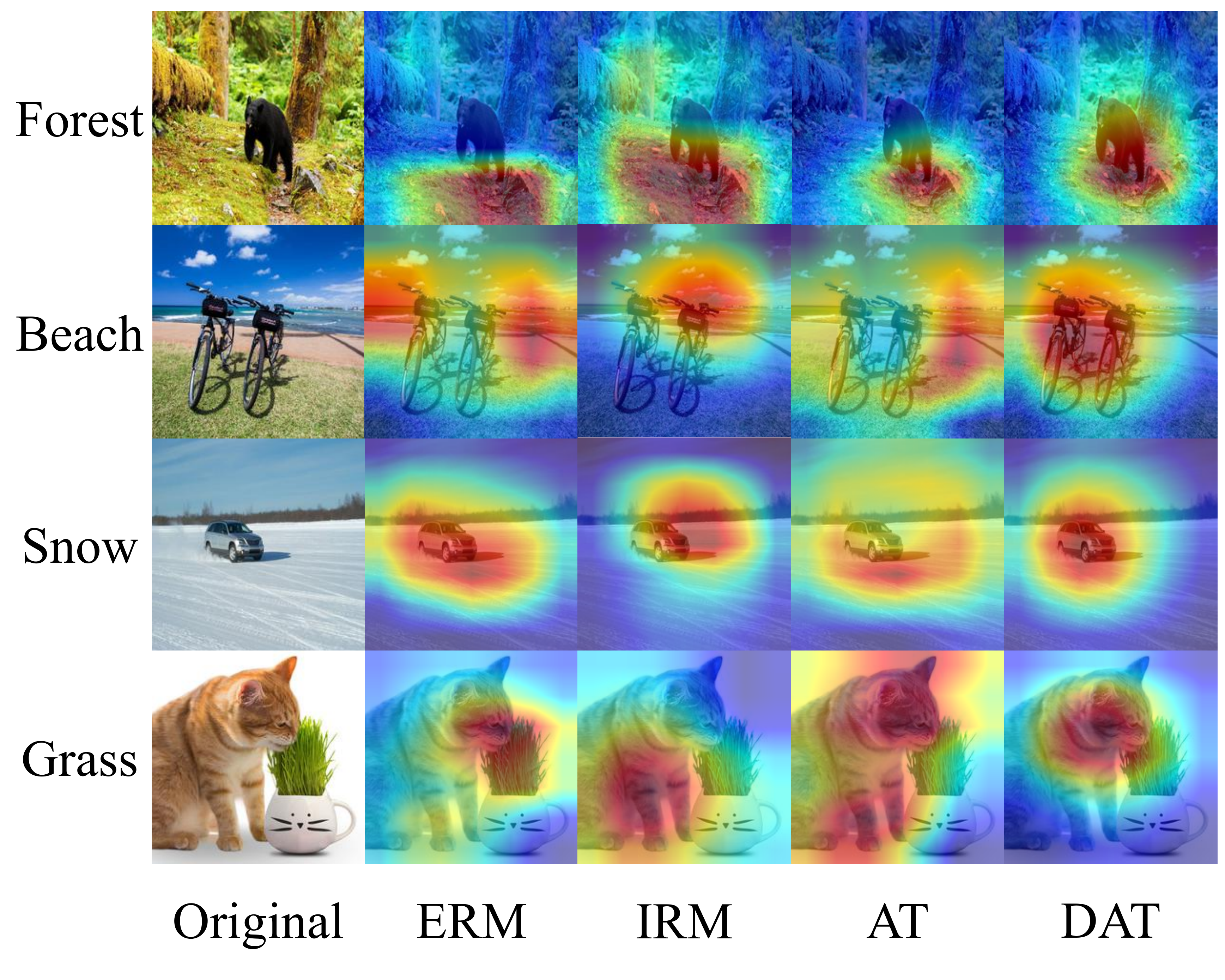}
    \caption{Attention heatmaps of ERM, IRM, (sample-wise) AT, and our DAT on the NICO dataset. Compared to other methods, DAT focuses more precisely on the object itself.}
    \label{fig:sem1}
\end{figure}

\paragraph{Qualitative Analysis through Semantic Graphs.} We use GradCam \cite{selvaraju2017grad, jacobgilpytorchcam} to visualize the attention heatmaps of models trained by ERM, sample-wise AT, and DAT on the NICO dataset. The results are shown in Figure \ref{fig:sem1}, where we can see that DAT pays more attention to the object itself than the strongly correlated background, while ERM and sample-wise AT tend to use environmental features instead.

\paragraph{Perturbation Radius and Step Size.} We investigate the effect of the perturbation radius $\varepsilon$ and the perturbation step size $\alpha$ on the NICO dataset. The results are shown in Table \ref{tab:EpsandAlpha}. The results show that the perturbation radius $\varepsilon$ greatly affects the OOD performance. When $\varepsilon$ is too large $(>10^{-1})$, it begins to hurt the invariant feature and causes performance degradation (from $72.9\%$ to $67.9\%$). The step size $\alpha$ has a smaller influence, and choosing a value between $1/100$ and $1/10$ times the size of $\varepsilon$ would be appropriate. The effect of the norm used for perturbation in DAT is analyzed in Appendix \ref{ablation}. 

\begin{table}[ht]

\centering
\begin{tabular}{ccc}
\toprule
Radius $\varepsilon$ & Step Size $\alpha$   & Acc (\%)  \\ \midrule
\multirow{3}{*}{$[10^{-2}, 10^{-1}]$} & $[10^{-4}, 10^{-3}]$ & $72.6 \pm 1.7$ \\
  & $[10^{-3}, 10^{-2}]$ & $72.0 \pm 2.1$ \\
  & $[10^{-2}, 10^{-1}]$ & $68.9 \pm 1.5$ \\
\midrule                     
\multirow{3}{*}{$[10^{-1}, 10^{0}]$}  & $[10^{-3}, 10^{-2}]$ & $71.2 \pm 0.4$ \\  
                                      & $[10^{-2}, 10^{-1}]$ & $66.6 \pm 1.7$ \\  
                     & $[10^{-1}, 10^{0}]$ & $69.4 \pm 0.4$\\
\midrule
\multirow{3}{*}{$[10^{0}, 10^{1}]$}   
& $[10^{-3}, 10^{-2}]$ & $64.6 \pm 0.3$               \\
& $[10^{-2}, 10^{-1}]$ & $67.9 \pm 2.0$ \\
& $[10^{-1}, 10^{0}]$ & $67.4 \pm 0.4$ \\
\bottomrule
\end{tabular}
\vspace{-0.1in}
\caption{Comparison of the test accuracy of different perturbation radius $\varepsilon$ and step size $\alpha$ of $\ell_{2}$-norm bounded DAT on the NICO dataset.}
\label{tab:EpsandAlpha}
\end{table}

\section{Discussions}
\paragraph{Comparison with Sample-wise AT.} Previous works \cite{hendrycks2021many, Yi2021ImprovedOG} try to exploit \emph{sample-wise} AT as a data augmentation strategy to obtain higher OOD performance. However, the performance only improves when the distribution shift is dominated by diversity shift, \eg noise, and blurring. Otherwise, performance might be degraded, as shown in Table \ref{table:bench}. One possible explanation is that sample-wise AT fails to capture the domain-level variations as DAT. As a result, it may add perturbations to the invariant features and hurt performance, especially under correlation shift.

\paragraph{Comparison with Invariant Causal Prediction.} A thread of methods, including ICP \cite{peters2016causal}, IRM \cite{arjovsky2019invariant}, and IGA \cite{koyama2020out}, try to find invariant data representations that could induce an invariant classifier. They have superior performance on synthetic datasets like CMNIST but fail to outperform ERM on real-world datasets (tasks dominated by both correlation shift and diversity shift). We believe that these failures could be attributed to the lack of prior information of their invariant learning principles. In our DAT, we effectively exploit the foreground-background difference in image classification tasks through domain-wise perturbations.

\section{Conclusion}
In this work, we carefully analyze the similarity between IRM and adversarial training in a domain-wise manner and establish a formal connection between OOD and adversarial robustness. Based on this connection, we propose a new adversarial training method for domain generalization: Domain-wise Adversarial Training (DAT). We show that it could effectively remove spurious background features in image classification and obtain fair performance on benchmark datasets. In particular, our DAT could consistently outperform ERM on tasks dominated by both the correlation shift and the diversity shift, while previous methods typically fail in one of the two cases.

\section*{Acknowledgment}
Yisen Wang is partially supported by the NSF China (No. 62006153), Open Research Projects of Zhejiang Lab (No. 2022RC0AB05), and Huawei Technologies Inc.

\bibliography{aaai23}

\newpage
\appendix
\onecolumn

\section{Experiment Settings and More Detailed Results}
\label{exp}

We train all the models based on the DomainBed benchmark presented in \cite{gulrajani2020search}. 

\subsection{Mixed-Same and Mixed-Rand}
\label{mixed}
To construct the dataset, we use the validation set of Mixed-Same (4185 images) provided by \cite{xiao2020noise} as the training environment and use the validation set of Mixed-Rand (4185 images) as the testing (OOD) environment. 

For training, we use pre-trained ResNet-18 as the backbone and train it for 5000 iterations. The hyperparameters are the same as those proposed in \cite{gulrajani2020search}, except that the learning rate searching interval is altered to $[10^{-5.5}, 10^{-4.5}]$, while we set the hyperparameters search space of DAT as $\varepsilon \in [10^{-0.5}, 10^{0}]$, $\alpha \in [10^{-1.5}, 10^{-1}]$.

\subsection{Typical OOD Datasets}
\label{dataset}
We conduct experiments on four typical OOD datasets, including CMNIST, NICO, Terra Incognita, and PACS. We split the NICO dataset using the same strategy in \cite{ye2021ood} for a fair comparison. 

For network architecture, models trained on CMNIST adopt the four-layer convolution network MNIST-CNN provided in the DomainBed benchmark \cite{gulrajani2020search}, while other datasets use ResNet-18 as the backbone. For Terra Incognita and PACS, pre-trained ResNet-18 is used. While on NICO, we use unpretrained ResNet-18 as the pre-training dataset contains photos that are largely overlapped with ImageNet classes.

For model selection, models trained on PACS and Terra Incognita are selected by training-domain validation, which selects the model with the highest training accuracy averaged across all training domains. For NICO, an additional OOD validation set is used. CMNIST, due to the large correlation shift, uses test-domain validation as its model selection criteria.

DAT, AT, and UAT models are trained for an appropriate number of iterations to ensure convergence, that is, $5000$ for Terra Incognita, $8000$ for NICO, and $10000$ for CMNIST and PACS.

When training NICO using DAT, we clamp the single sample adversarial loss to $(0,2)$ to obtain a better domain-wise perturbation as suggested in \cite{shafahi2020universal}.

We use the same hyperparameters as those proposed in \cite{ye2021ood} whenever possible while setting the hyperparameters search space for DAT, sample-wise AT, and UAT as in Table \ref{tab:searchd}, \ref{tab:searcha} and \ref{tab:searchu}. We use DAT with perturbation bounded by $\ell_2$-norm in our experiments. For sample-wise AT, we use a 10-step $\ell_2$ PGD \cite{madry2018towards}.

For WRM, we set the step number to $10$ in inner maximization while searching for the learning rate in inner maximization in the range $[10^{-2}, 10^{-1}]$ and $\gamma$ in the range $[10^{-0.5}, 10^{0.5}]$. We train the models for $5000$ iterations to ensure convergence.

For ADA, we set the step number to $10$ in the inner maximization while searching for the adversarial learning rate in the range $[10^{-1}, 10^{0.5}]$, $\gamma$ in the range $[0.5, 1.5]$, the number of steps in the min-phase in the range $[10^{0.5}, 10^{2.5}]$, and the number of whole adversarial phases in the range $[10^{0.5}, 10^{2}]$. We train the models for $5000$ iterations to ensure convergence.

\begin{table}[H]
    \centering
    \begin{tabular}{ccc}
\toprule
Dataset & Radius $\varepsilon$ & Step Size $\alpha$   \\ \midrule
CMNIST & $[10^{-1}, 10^{2}]$               & $[10^{-2}, 10^{1}]$ \\ 
NICO & $[10^{-2}, 10^{-1}]$               & $[10^{-4}, 10^{-3}]$\\ 
PACS & $[10^{-1}, 10^{2}]$                & $[10^{-2}, 10^{1}]$\\ 
TerraInc & $[10^{-0.5}, 10^{0.7}]$                & $[10^{-2}, 10^{-1}]$\\
\bottomrule
\end{tabular}
\caption{Hyperparameter Search Space for DAT on Typical OOD Datasets.}
\label{tab:searchd}
\end{table}

\begin{table}[H]
    \centering
    \begin{tabular}{ccc}
\toprule
Dataset & Radius $\varepsilon$ & Step Size $\alpha$   \\ \midrule
CMNIST & $[10^{-1}, 10^{1}]$               & $[10^{-2}, 10^{0}]$ \\ 
NICO & $[10^{-1}, 10^{0}]$               & $[10^{-2}, 10^{-1}]$\\ 
PACS & $[10^{-2}, 10^{-1}]$                & $[10^{-3}, 10^{-2}]$\\ 
TerraInc & $[10^{-1}, 10^{0}]$                & $[10^{-2}, 10^{-1}]$\\
\bottomrule
\end{tabular}
\caption{Hyperparameter Search Space for (sample-wise) AT on Typical OOD Datasets.}
    \label{tab:searcha}
\end{table}

\begin{table}[H]
    \centering
    \begin{tabular}{ccc}
\toprule
Dataset & Radius $\varepsilon$ & Step Size $\alpha$   \\ \midrule
CMNIST & $[10^{-1}, 10^{2}]$               & $[10^{-2}, 10^{1}]$ \\ 
NICO & $[10^{-2}, 10^{-1}]$               & $[10^{-4}, 10^{-3}]$\\ 
PACS & $[10^{-1}, 10^{2}]$                & $[10^{-2}, 10^{1}]$\\ 
TerraInc & $[10^{0}, 10^{0.5}]$                & $[10^{-2}, 10^{-1}]$\\
\bottomrule
\end{tabular}
\caption{Hyperparameter Search Space for UAT on Typical OOD Datasets.}
\label{tab:searchu}
\end{table}

The domain-split results for ERM, UAT, and DAT are shown in Table \ref{tab:cmnist}, \ref{tab:pacs}, and \ref{tab:terrain}, where the results of ERM on CMNIST are from \cite{gulrajani2020search} and all other results come from our experiments.

\begin{table}[H]
\begin{subtable}[h]{1\textwidth}
\centering
\caption{Training domain validation.}
\begin{tabular}{clllll}
\toprule
\multicolumn{1}{l}{Algorithm} &  $0.1$          & $0.2$          & $0.9$          & \text{Average} \\ \midrule
{ERM}   & $\mathbf{72.7 \pm 0.2}$ & $73.2 \pm 0.3$ & $10.0 \pm 0.0$ & $52.0 \pm 0.1$ \\
{UAT}   & $72.1 \pm 0.1$ & $\mathbf{73.7 \pm 0.1}$ & $10.0 \pm 0.1$ & $52.0 \pm 0.1$ \\
{DAT}   & $72.4 \pm 0.2$ & $\mathbf{73.7 \pm 0.1}$ & $\mathbf{10.2 \pm 0.2}$ & $\mathbf{52.1 \pm 0.1}$ \\ \bottomrule
\end{tabular}
\end{subtable}
\begin{subtable}[h]{\textwidth}
\centering
\caption{Test domain validation (Oracle).}
\begin{tabular}{clllll}
\toprule
\multicolumn{1}{l}{Algorithm} &  $0.1$          & $0.2$          & $0.9$          & \text{Average} \\ \midrule
{ERM}   & $72.3 \pm 0.6$ & $73.1 \pm 0.3$ & $30.0 \pm 0.3$ & $58.5 \pm 0.3$ \\
{UAT}   & $75.3 \pm 6.4$ & $71.9 \pm 0.7$ & $29.0 \pm 0.2$ & $58.7 \pm 2.3$ \\
{DAT}   & $\mathbf{78.5 \pm 4.8}$ & $\mathbf{79.3 \pm 0.2}$ & $\mathbf{47.5 \pm 2.6}$ & $\mathbf{68.4 \pm 2.0}$ \\ \bottomrule
\end{tabular}
\end{subtable}
\caption{Test accuracy of ERM, UAT, and DAT on CMNIST.}
\label{tab:cmnist}
\end{table}

\begin{table}[H]
\begin{subtable}[h]{\textwidth}
\centering
\begin{tabular}{cllllll}
\toprule
\multicolumn{1}{l}{Algorithm} &  Artpaint & Cartoon              & Photo              & Sketch  & Average    \\ \midrule
{ERM}   & $\mathbf{80.5 \pm 0.8}$ & $74.2 \pm 0.5$ & $\mathbf{94.7 \pm 0.5}$ & $72.9 \pm 2.1$ & $80.6 \pm 0.6$ \\
{UAT}   & $79.9 \pm 0.6$ & $74.7 \pm 0.8$ & $92.6 \pm 0.4$ & $75.5 \pm 2.1$ & $80.7 \pm 0.4$ \\
{DAT}   & $80.0 \pm 0.3$ & $\mathbf{77.6 \pm 0.8}$ & $92.6 \pm 0.9$ & $\mathbf{77.6 \pm 0.5}$ & $\mathbf{82.0 \pm 0.1}$ \\ \bottomrule
\end{tabular}
\end{subtable}
\begin{subtable}[h]{\textwidth}
\centering
\caption{Test domain validation (Oracle).}
\begin{tabular}{cllllll}
\toprule
\multicolumn{1}{l}{Algorithm} &  Artpaint & Cartoon              & Photo              & Sketch  & Average    \\ \midrule
{ERM}   & $79.7 \pm 1.0$ & $\mathbf{76.9 \pm 0.3}$ & $\mathbf{94.2 \pm 0.4}$ & $78.3 \pm 1.5$ & $\mathbf{82.3 \pm 0.4}$ \\
{UAT}   & $\mathbf{80.4 \pm 2.1}$ & $75.0 \pm 0.4$ & $93.5 \pm 0.5$ & $\mathbf{79.3 \pm 0.4}$ & $82.0 \pm 0.6$ \\
{DAT}   & $79.7 \pm 0.1$ & $74.9 \pm 0.7$ & $93.9 \pm 0.6$ & $78.1 \pm 0.4$ & $81.7 \pm 0.2$ \\ \bottomrule
\end{tabular}
\caption{Training domain validation.}
\end{subtable}
\caption{Test accuracy of ERM, UAT, and DAT on PACS.}
\label{tab:pacs}
\end{table}

\begin{table}[H]
\begin{subtable}[h]{\textwidth}
\centering
\begin{tabular}{cllllll}
\toprule
\multicolumn{1}{l}{Algorithm} &  L100           & L38            & L43            & L46  & Average    \\ \midrule
{ERM}   & $53.2 \pm 1.4$ & $31.9 \pm 0.3$ & $50.2 \pm 0.9$ & $\mathbf{33.6 \pm 0.3}$ & $42.2 \pm 0.4$ \\
{UAT}   & $47.2 \pm 3.3$ & $\mathbf{36.9 \pm 1.6}$ & $\mathbf{50.3 \pm 0.3}$ & $32.8 \pm 0.3$ & $41.8 \pm 1.2$ \\
{DAT}   & $\mathbf{53.8 \pm 1.6}$ & $35.1 \pm 2.3$ & $\mathbf{50.3 \pm 0.3}$ & $31.7 \pm 0.6$ & $\mathbf{42.7 \pm 0.7}$ \\ \bottomrule
\end{tabular}
\caption{Training domain validation.}
\end{subtable}
\begin{subtable}[h]{\textwidth}
\centering

\begin{tabular}{cllllll}
\toprule
\multicolumn{1}{l}{Algorithm} &  L100           & L38            & L43            & L46  & Average    \\ \midrule
{ERM}   & $\mathbf{55.8 \pm 1.1}$ & $40.8 \pm 2.0$ & $49.8 \pm 1.3$ & $35.0 \pm 0.2$ & $45.3 \pm 0.3$ \\
{UAT}   & $50.5 \pm 2.6$ & $40.3 \pm 0.2$ & $50.7 \pm 0.3$ & $34.3 \pm 0.5$ & $44.0 \pm 0.6$ \\
{DAT}   & $54.1 \pm 2.3$ & $\mathbf{42.5 \pm 2.1}$ & $\mathbf{52.7 \pm 0.8}$ & $\mathbf{35.1 \pm 1.0}$ & $\mathbf{46.1 \pm 0.6}$ \\ \bottomrule
\end{tabular}
\caption{Test domain validation (Oracle).}
\end{subtable}
\caption{Test accuracy of ERM, UAT, and DAT on TerraInc.}
\label{tab:terrain}
\end{table}

\subsection{DomainBed}
In this section, we present the performance of DAT on DomainBed benchmark \cite{gulrajani2020search} except DomainNet in Table \ref{tab:db}. The results for all other baselines are from the official repository of DomainBed. We highlight the results where DAT performs better than ERM. From the results, we can see that DAT outperforms ERM on most datasets while having comparable performance on other datasets.
\label{domainbed}

\begin{table}[ht]
\centering
\begin{tabular}{lccccccc}
\toprule
\textbf{Algorithm}        & \textbf{ColoredMNIST}     & \textbf{RotatedMNIST}     & \textbf{VLCS}             & \textbf{PACS}             & \textbf{OfficeHome}       & \textbf{TerraIncognita}   &\textbf{Avg}              \\
\midrule
ERM                       & 51.5 $\pm$ 0.1            & 98.0 $\pm$ 0.0            & 77.5 $\pm$ 0.4            & 85.5 $\pm$ 0.2            & 66.5 $\pm$ 0.3            & 46.1 $\pm$ 1.8                  & 70.9                    \\\midrule
IRM                       & 52.0 $\pm$ 0.1            & 97.7 $\pm$ 0.1            & 78.5 $\pm$ 0.5            & 83.5 $\pm$ 0.8            & 64.3 $\pm$ 2.2            & 47.6 $\pm$ 0.8                   & 70.6                    \\
GroupDRO                  & 52.1 $\pm$ 0.0            & 98.0 $\pm$ 0.0            & 76.7 $\pm$ 0.6            & 84.4 $\pm$ 0.8            & 66.0 $\pm$ 0.7            & 43.2 $\pm$ 1.1                    & 70.1                   \\
Mixup                     & 52.1 $\pm$ 0.2            & 98.0 $\pm$ 0.1            & 77.4 $\pm$ 0.6            & 84.6 $\pm$ 0.6            & 68.1 $\pm$ 0.3            & 47.9 $\pm$ 0.8                     & 71.3                   \\
MLDG                      & 51.5 $\pm$ 0.1            & 97.9 $\pm$ 0.0            & 77.2 $\pm$ 0.4            & 84.9 $\pm$ 1.0            & 66.8 $\pm$ 0.6            & 47.7 $\pm$ 0.9                  & 71.0                  \\
CORAL                     & 51.5 $\pm$ 0.1            & 98.0 $\pm$ 0.1            & 78.8 $\pm$ 0.6            & 86.2 $\pm$ 0.3            & 68.7 $\pm$ 0.3            & 47.6 $\pm$ 1.0            & 71.8                   \\
MMD                       & 51.5 $\pm$ 0.2            & 97.9 $\pm$ 0.0            & 77.5 $\pm$ 0.9            & 84.6 $\pm$ 0.5            & 66.3 $\pm$ 0.1            & 42.2 $\pm$ 1.6                  & 70.0                 \\
DANN                      & 51.5 $\pm$ 0.3            & 97.8 $\pm$ 0.1            & 78.6 $\pm$ 0.4            & 83.6 $\pm$ 0.4            & 65.9 $\pm$ 0.6            & 46.7 $\pm$ 0.5              & 70.7                  \\
CDANN                     & 51.7 $\pm$ 0.1            & 97.9 $\pm$ 0.1            & 77.5 $\pm$ 0.1            & 82.6 $\pm$ 0.9            & 65.8 $\pm$ 1.3            & 45.8 $\pm$ 1.6                    & 70.2               \\
MTL                       & 51.4 $\pm$ 0.1            & 97.9 $\pm$ 0.0            & 77.2 $\pm$ 0.4            & 84.6 $\pm$ 0.5            & 66.4 $\pm$ 0.5            & 45.6 $\pm$ 1.2                   & 70.5                    \\
SagNet                    & 51.7 $\pm$ 0.0            & 98.0 $\pm$ 0.0            & 77.8 $\pm$ 0.5            & 86.3 $\pm$ 0.2            & 68.1 $\pm$ 0.1            & 48.6 $\pm$ 1.0                  & 71.7                 \\
ARM                       & 56.2 $\pm$ 0.2            & 98.2 $\pm$ 0.1            & 77.6 $\pm$ 0.3            & 85.1 $\pm$ 0.4            & 64.8 $\pm$ 0.3            & 45.5 $\pm$ 0.3                & 71.2                    \\
VREx                      & 51.8 $\pm$ 0.1            & 97.9 $\pm$ 0.1            & 78.3 $\pm$ 0.2            & 84.9 $\pm$ 0.6            & 66.4 $\pm$ 0.6            & 46.4 $\pm$ 0.6           & 70.9                      \\
RSC                       & 51.7 $\pm$ 0.2            & 97.6 $\pm$ 0.1            & 77.1 $\pm$ 0.5            & 85.2 $\pm$ 0.9            & 65.5 $\pm$ 0.9            & 46.6 $\pm$ 1.0                   & 70.6              \\
\midrule
DAT & \textbf{52.1 $\pm$ 0.1} & 97.9 $\pm$ 0.0 & 77.2 $\pm$ 0.3 & \textbf{86.0 $\pm$ 0.4} &  \textbf{67.0 $\pm$ 0.4} & \textbf{47.1 $\pm$ 0.8} & \textbf{71.2}\\
\bottomrule
\end{tabular}
\caption{Sweep results on DomainBed.}
\label{tab:db}
\end{table}

\subsection{The Effect of the Norm Used for Perturbation in DAT}
\label{ablation}

In this section, we provide the results of DAT on the NICO dataset using $\ell_\infty$ as the perturbation norm instead of $\ell_2$ used in Table \ref{tab:EpsandAlpha} in Section \ref{analysis}. The results are shown in Table \ref{tab:inf}, which shows that we could still obtain a good performance in this scenario by choosing a sufficiently large perturbation radius.

\begin{table}[H]
\centering
\begin{tabular}{ccc}
\toprule
Radius $\varepsilon$ & Step Size $\alpha$   & Acc (\%)  \\ \midrule
$[10^{-3}, 10^{-2}]$               & $[10^{-4}, 10^{-3}]$ & $68.5 \pm 0.2$ \\ 
$[10^{-2}, 10^{-1}]$               & $[10^{-3}, 10^{-2}]$ & $67.9 \pm 2.4$ \\ 
$[10^{-1}, 10^{0}]$                & $[10^{-3}, 10^{-2}]$ & $72.5 \pm 1.6$ \\ 
\bottomrule
\end{tabular}
\caption{Test accuracy of $\ell_{\infty}$-norm bounded DAT on the NICO dataset.}
\label{tab:inf}
\end{table}

\subsection{The Effect of the Batch Size Used for Perturbation in DAT}
\begin{figure}[t]
    \centering
    \includegraphics[width=0.4\linewidth]{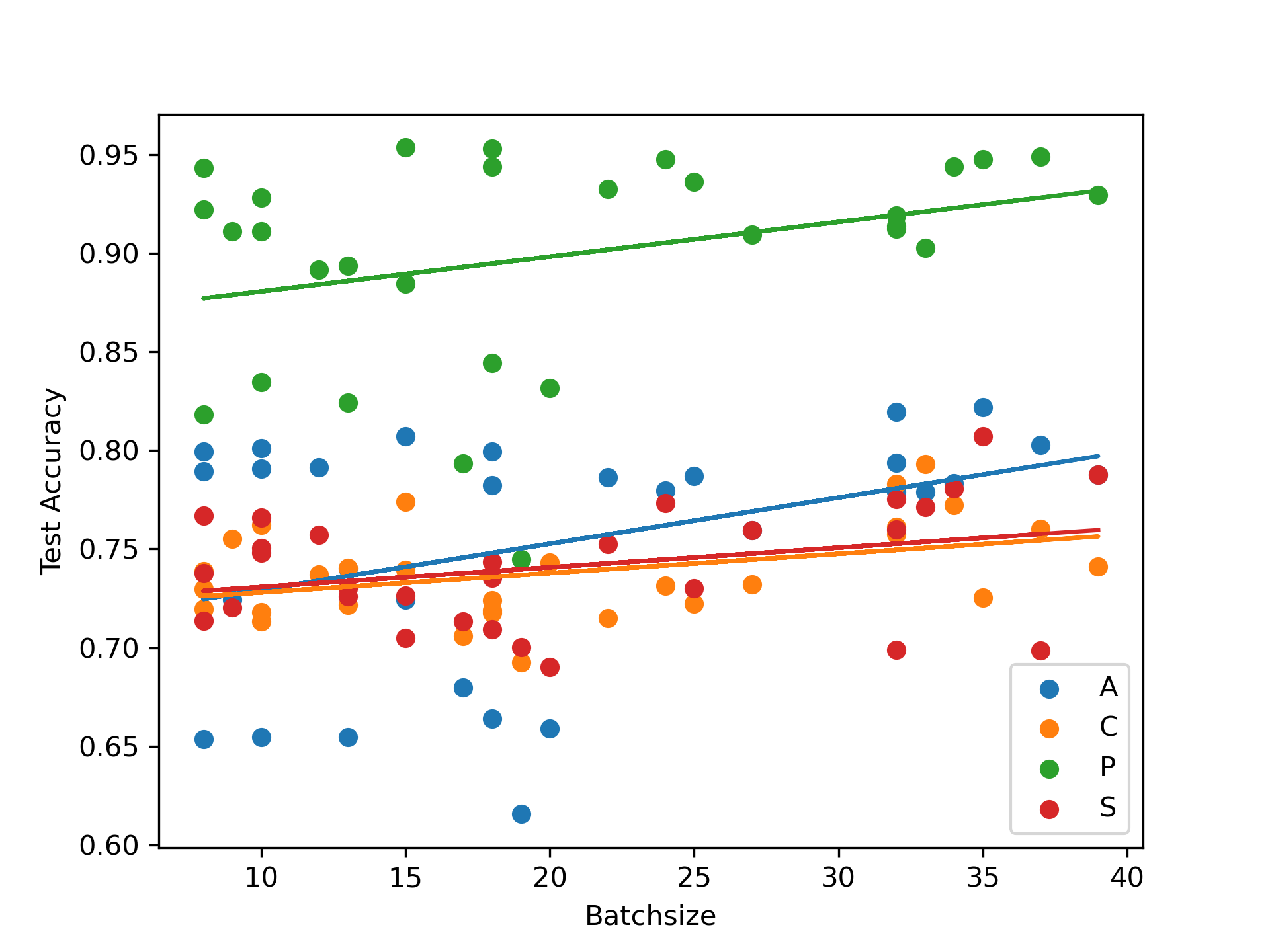}
    \caption{The Test Accuracy of DAT models with different batch size on the four domains of PACS. Regression lines are shown.}
    \label{fig:batchsize}
\end{figure}

As DAT calculates the increment of the perturbation based on a batch of inputs, the performance correlates with the batch size during training. As the batch size we used is sampled in a search space, we empirically verified this by showing the scatter plot of the batch size and the test accuracy on PACS, as shown in Fig. \ref{fig:batchsize}. We can see that the batch size has a positive correlation with the performance, which means that a better approximation of the distributional level update of the perturbation in each step leads to better performance, which does support our underlying hypothesis that generating adversarial perturbations with respect to domain distribution could lead to better OOD generalization performance.

\section{On Single-source Domain Generalization}

\label{ssdg}
To show that our method can also be applied to single-source domain generalization scenarios, we conduct experiments with settings similar to those of Appendix \ref{dataset} on the four benchmark datasets, except that we train on one of the domains and report its test accuracy on all other domains. For the NICO dataset, as it comes with a validation domain, we train the models on one of the training domains, using the validation domain for model selection, then report its accuracy on the test domain.

We train all models using ERM/DAT for $5000$ iterations to ensure convergence and set the hyperparameter search space of DAT as shown in Table \ref{tab:searchsingle}. All other settings are the same as those shown in Appendix \ref{dataset}.

\begin{table}[H]
    \centering
    \begin{tabular}{ccc}
\toprule
Dataset & Radius $\varepsilon$ & Step Size $\alpha$   \\ \midrule
CMNIST & $[10^{-1}, 10^{2}]$               & $[10^{-2}, 10^{1}]$ \\ 
NICO & $[10^{-2}, 10^{-1}]$               & $[10^{-4}, 10^{-3}]$\\ 
PACS & $[10^{-2}, 10^{-1}]$                & $[10^{-3}, 10^{-2}]$\\ 
TerraInc & $[10^{-0.5}, 10^{0.7}]$                & $[10^{-2}, 10^{-1}]$\\
\bottomrule
\end{tabular}
\caption{Hyperparameter searching space for DAT on typical OOD datasets (single-source domain generalization).}
    \label{tab:searchsingle}
\end{table}

For the Digits dataset, we used the exact same setting as in \cite{Qiao2020LearningTL}, where the model is trained on the MNIST dataset and tested on all other four datasets.

The domain-split results for ERM and DAT under single-source domain generalization setting are shown in Table \ref{tab:cmnistes} to \ref{tab:terrads}.
\begin{table}[H]
\begin{subtable}[h] {\textwidth}
\centering
\begin{tabular}{cllll}
\toprule
\multicolumn{1}{l}{Training Domain} & 0.1          & 0.2          & 0.9          & Avg  \\ \midrule
0.1             & \textbackslash{}            & $79.8 \pm 0.2$ & $21.2 \pm 0.2$ & $50.5$ \\
0.2             & $88.7 \pm 0.4$ & \textbackslash{}            & $35.2 \pm 1.3$ & $\mathbf{62.0}$ \\
0.9             & $21.0 \pm 0.9$ & $29.1 \pm 0.7$ & \textbackslash{}            & $\mathbf{25.1}$ \\ \bottomrule
\end{tabular}
\caption{ERM on CMNIST.}
\label{tab:cmnistes}
\end{subtable}

\begin{subtable}[h] {\textwidth}
\centering
\begin{tabular}{cllll}
\toprule
\multicolumn{1}{l}{Training Domain} & 0.1              & 0.2              & 0.9              & Avg  \\ \midrule
0.1             & \textbackslash{} & $80.0 \pm 0.1$     & $27.9 \pm 4.7$     & $\mathbf{54.0}$ \\
0.2             & $90.0 \pm 0.1$     & \textbackslash{} & $32.4 \pm 1.0$     & $61.2$ \\
0.9             & $18.3 \pm 1.1$     & $27.1 \pm 1.1$     & \textbackslash{} & $22.7$ \\ \bottomrule
\end{tabular}
\caption{DAT on CMNIST.}
\label{tab:cmnistds}
\end{subtable}

\begin{subtable}[h] {\textwidth}
\centering
\begin{tabular}{cl}
\toprule
\multicolumn{1}{l}{Training Domain} & Test Accuracy \\ \midrule
1               & $56.8 \pm 3.5$  \\
2               & $69.6 \pm 2.0$  \\
Avg             & $63.2$         \\ \bottomrule
\end{tabular}
\caption{ERM on NICO.}
\label{tab:nicoes}
\end{subtable}

\begin{subtable}[h] {\textwidth}
\centering
\begin{tabular}{cl}
\toprule
\multicolumn{1}{l}{Training Domain} & Test Accuracy \\ \midrule
1               & $61.7 \pm 0.9$  \\
2               & $67.1 \pm 0.3$  \\
Avg             & $\mathbf{64.4}$         \\ \bottomrule
\end{tabular}
\caption{DAT on NICO.}
\label{tab:nicods}
\end{subtable}

\begin{subtable}[h] {\textwidth}
\centering
\begin{tabular}{clllll}
\toprule
\multicolumn{1}{l}{Training Domain} & Artpaint         & Cartoon          & Photo            & Sketch           & Avg  \\ \midrule
Artpaint        & \textbackslash{} & $64.9 \pm 0.7$     & $94.2 \pm 0.1$     & $61.5 \pm 4.1$     & $\mathbf{73.5}$ \\
Cartoon         & $61.9 \pm 2.7$     & \textbackslash{} & $76.6 \pm 1.6$     & $69.5 \pm 1.1$     & $69.3$ \\
Photo           & $66.9 \pm 0.3$     & $26.9 \pm 0.1$     & \textbackslash{} & $35.5 \pm 0.7$     & $\mathbf{43.1}$ \\
Sketch          & $47.3 \pm 0.1$     & $65.7 \pm 0.9$     & $46.6 \pm 0.8$     & \textbackslash{} & $53.2$ \\ \bottomrule
\end{tabular}
\caption{ERM on PACS.}
\label{tab:pacses}
\end{subtable}

\begin{subtable}[h] {\textwidth}
\centering
\begin{tabular}{clllll}
\toprule
\multicolumn{1}{l}{Training Domain} & Artpaint         & Cartoon          & Photo            & Sketch           & Avg  \\ \midrule
Artpaint        & \textbackslash{} & $59.6 \pm 0.0$     & $94.2 \pm 0.8$     & $48.7 \pm 2.3$     & $67.5$ \\
Cartoon         & $68.1 \pm 0.1$     & \textbackslash{} & $83.9 \pm 1.0$     & $69.1 \pm 1.3$     & $\mathbf{73.7}$ \\
Photo           & $64.5 \pm 1.4$     & $27.6 \pm 5.8$     & \textbackslash{} & $30.9 \pm 4.5$     & $41.0$ \\
Sketch          & $52.2 \pm 0.2$     & $63.4 \pm 1.4$     & $56.4 \pm 1.1$     & \textbackslash{} & $\mathbf{57.3}$ \\ \bottomrule
\end{tabular}
\caption{DAT on PACS.}
\label{tab:pacsds}
\end{subtable}

\begin{subtable}[h] {\textwidth}
\centering
\begin{tabular}{clllll}
\toprule
\multicolumn{1}{l}{Training Domain} & L100             & L38              & L43              & L46              & Avg  \\ \midrule
L100            & \textbackslash{} & $24.9 \pm 4.1$     & $26.8 \pm 0.5 $    & $27.1 \pm 4.5$     & $\mathbf{26.3}$ \\
L38             & $42.3 \pm 0.6$     & \textbackslash{} & $12.5 \pm 1.2$     & $13.2 \pm 1.5$     & $22.7$ \\
L43             & $34.2 \pm 9.8$     & $31.9 \pm 0.9$     & \textbackslash{} & $30.1 \pm 0.5$     & $32.1$ \\
L46             & $24.9 \pm 6.2$     & $13.0 \pm 2.8$     & $46.6 \pm 2.0$     & \textbackslash{} & $28.3$ \\ \bottomrule
\end{tabular}
\caption{ERM on TerraInc.}
\label{tab:terraes}
\end{subtable}

\begin{subtable}[h] {\textwidth}
\centering
\begin{tabular}{clllll}
\toprule
\multicolumn{1}{l}{Training Domain} & L100             & L38              & L43              & L46              & Avg  \\ \midrule
L100                                  & \textbackslash{} & $23.3 \pm 1.4$     & $24.3 \pm 0.7$     & $16.7 \pm 1.2$     & $21.4$ \\
L38                                   & $44.6 \pm 7.4$     & \textbackslash{} & $17.3 \pm 1.0$     & $16.3 \pm 0.4$     & $\mathbf{26.1}$ \\
L43                                   & $35.2 \pm 7.4$     & $32.0 \pm 7.6$     & \textbackslash{} & $32.9 \pm 1.$5     & $\mathbf{33.4}$ \\
L46                                   & $24.3 \pm 4.0$     & $23.2 \pm 0.2$     & $47.5 \pm 0.2$     & \textbackslash{} & $\mathbf{31.7}$ \\ \bottomrule
\end{tabular}
\caption{DAT on TerraInc.}
\label{tab:terrads}
\end{subtable}
\caption{Test accuracy (single-source domain generalization).}
\end{table}

\section{Proofs}
\label{proof}

We give a proof of Remark \ref{remarka} (Remark \ref{remark}) first, then use a similar technique to prove Proposition \ref{prop2a} (Proposition \ref{prop2}).

\begin{remark}[Equivalence under Single-sample Environments]
\label{remarka}
    When the environments degenerate into a single data point, we have the following relationship:
If $\varepsilon$ is sufficiently small, then for $\beta\cdot\Phi$ as a deep network with any activation function,
the penalty term of IRMv1 (Eq. \ref{IRMv1Obj}) on each sample and the square of the maximization term of Linearized version of Eq.~\ref{eq:dat} (LDAT, obtained by first-order approximation of DAT) 
\begin{equation}
    \mathrm{Penalty_{LDAT}} = \left\langle\nabla_{x} \ell\left(\beta^{T} \Phi(x), y\right), \pm\hat\delta_x\right\rangle
\end{equation}
on each sample with perturbation\ $\hat\delta_x = \pm\varepsilon x$ only differ by a fixed multiple $\varepsilon^2$ and a bias term $B_x$, which is formally stated as
\begin{align}
\mathrm{Penalty^{2}_{LDAT}}
&= [\langle \nabla_{x} \ell\left(\beta^{T} \Phi(x), y\right),\pm \varepsilon x\rangle]^2 =\varepsilon^2 (1-\sigma(\beta^\top\Phi(x)))^2 \left\|\beta^\top \Phi_x x\right\|^2\\
&= \varepsilon^{2}\cdot\mathrm{Penalty^\prime_{IRM}}.
\end{align}
\end{remark}
where $\mathrm{Penalty^\prime_{IRM}}=(1-\sigma(y\beta^\top\Phi(x)))^2 \left\|\beta^\top(\Phi_x x)\right\|^2$ and $\mathrm{Penalty_{IRM}}=(1-\sigma(y\beta^\top\Phi(x)))^2 \left\|\beta^\top(\Phi_x x+B_x)\right\|^2$.

\paragraph{Proof of Remark \ref{remarka}}

Note that we presume $\Phi$ is piece-wise linear. We represent the output logit of the given deep network as $\beta^\top\Phi(x)$ = $\beta^\top (\Phi_x x + B_x)$ where $\Phi_x$ and $B_x$ are related to the sample $x$ due to the different activation patterns.

We first derive the form of $\mathrm{Penalty_{LDAT}}$ by first-order approximation of Eq.~\ref{eq:dat}.
\begin{align}
    \ell(\beta^\top \Phi(x+\delta_e),y)-\ell(\beta^\top \Phi(x),y) \approx \left\langle \nabla_x \ell(\beta^\top\Phi(x),y), \delta_e\right\rangle
\end{align}
Letting $\delta_e = \pm\varepsilon x$, we have
\begin{equation}
    \mathrm{Penalty_{LDAT}} = \left\langle\nabla_{x} \ell\left(\beta^{T} \Phi(x), y\right), \pm\hat\delta_x\right\rangle
\end{equation}

The penalty term for IRMv1 on each sample is as follows:
\begin{equation}
\label{IRMsample}
\begin{aligned}
    &\left\|\nabla_{w|w=1.0}-\log(\sigma(w\cdot(y \beta^\top\Phi(x))))\right\|^2\\
    &=\left\|-\frac{\sigma'(w\cdot(y\beta^\top\Phi(x)))}{\sigma(w\cdot(y\beta^\top\Phi(x)))} y\beta^\top \Phi(x)\right\|^2\\
    &=\left\|-(1-\sigma(y\beta^\top\Phi(x))) y\beta^\top\Phi(x)\right\|^2\\
    &=(1-\sigma(y\beta^\top\Phi(x)))^2 \left\|\beta^\top(\Phi_x x+B_x)\right\|^2
\end{aligned}
\end{equation}
For LDAT, note that the gradient \wrt~$x$ is equal to
$$
\nabla_x [-\log(\sigma(y \beta^\top\Phi(x))] = -(1-\sigma(y\beta^\top\Phi(x))) y\beta^\top \Phi_x
$$
So the square of the penalty term of LDAT on each sample with perturbation $\pm\varepsilon x$ is:
\begin{equation}
    [\langle \nabla_x -\log(\sigma(y \beta^\top\Phi(x))),\pm \varepsilon x\rangle]^2 =\varepsilon^2 (1-\sigma(\beta^\top\Phi(x)))^2 \left\|\beta^\top \Phi_x x\right\|^2
\end{equation}
which is identical to Eq.~\ref{IRMv1Obj} with a difference of multiple $\varepsilon^2$ and a bias term $B_x$.

\qedwhite

\begin{prop} \label{prop2a}
Consider each $D_e$ as  the corresponding distribution of a particular training domain $e$. For any $\beta\cdot\Phi$ as a deep network with any activation function, the penalty term of IRMv1, $\mathrm{Penalty_{IRM}}$ (Eq.~\ref{IRMv1Obj}), could be expressed as the square of a reweighted version of the penalty term of the above approximate target, $\mathrm{Penalty_{DAT}}$(Eq.~\ref{eq:dat2}), on each environment $e$ with coefficients related to 
the distribution $D_e$, which could be stated as follows:
\begin{align}
    \mathrm{Penalty_{IRM}} 
    &= \left\|\mathbb{E}_{D_{e}} [L_x x + \tilde{B}_x]\right\|^2\\ %
    \mathrm{Penalty_{DAT}} &= \left\|\mathbb{E}_{D_{e}} L_x\right\|
\end{align}
where $L_x = (1-\sigma(y\beta^\top\Phi_x x))y\beta^\top\Phi_x$ and $\tilde{B}_{x}=\left(1-\sigma\left(y \beta^{\top} \Phi_{x} x\right)\right) y \beta^{\top} B_{x}$. $B_x$ denotes the collection of constants (introduced by bias terms in neural network layers).
\end{prop}

\paragraph{Proof of Proposition \ref{prop2a}}

For IRMv1, from the proof of Remark \ref{remarka} we know
\begin{equation}
\nabla_{w|w=1.0} \ell(w\cdot(\beta\cdot \Phi),y) = -(1-\sigma(y\beta^\top\Phi_x x))y\beta^\top(\Phi_x x + B_x)
\end{equation}
So that $\mathrm{Penalty_{IRM}}$ (suppose derivation and integration are commutable)
\begin{equation}
\begin{aligned}
    \left\|\nabla_{w|w=1.0} R^e(w\cdot(\beta\cdot\Phi))\right\|^2
    &=\left\|\mathbb E_{D_e} (1-\sigma(y\beta^\top\Phi_x x))y\beta^\top(\Phi_x x + B_x)\right\|^2\\
\end{aligned}
\end{equation}
For DAT, on each environment

\begin{equation}
    \begin{aligned}
    &\max_{\|\delta_e\|\le \varepsilon}\mathbb E_{D_e}\left[ \ell(\beta^\top \Phi(x+\delta_e),y)-\ell(\beta^\top \Phi(x),y)\right)\big]\\
\approx &\max_{\|\delta_e\|\le \varepsilon}\mathbb E_{D_e}\left[ \left\langle \nabla_x\ell(\beta^\top \Phi(x),y), \delta_e\right\rangle\right]\\
= &\max_{\|\delta_e\|\le \varepsilon}\left\langle\mathbb E_{D_e}\left[\nabla_x\ell(\beta^\top \Phi(x),y) \right], \delta_e\right\rangle\\
= &\varepsilon\left\|\mathbb E_{ D_e}\left[\nabla_x \ell(\beta^\top\Phi(x),y)\right]\right\|\\
\end{aligned}
\end{equation}
So $\mathrm{Penalty_{DAT}}$ is
\begin{equation}
    \left\|\mathbb E_{ D_e}\left[\nabla_x \ell(\beta^\top\Phi(x),y)\right]\right\|
= \left\|\mathbb E_{D_e} \left[-(1-\sigma(y\beta^\top\Phi(x))) y\beta^\top \Phi_x\right]\right\|
\end{equation}
Therefore, we know that $\mathrm{Penalty_{IRM}}$ can be regarded as the square of a \textit{reweighted} version of $\mathrm{Penalty_{DAT}}$ with the coefficient on each sample within the expectation $x$ with the difference of a bias term $B_x$.

\qedwhite

\end{document}